\documentclass{article}
\usepackage{ijcai22}

\usepackage{times}
\usepackage{soul}
\usepackage{url}
\usepackage[hidelinks]{hyperref}
\usepackage[utf8]{inputenc}
\usepackage[small]{caption}
\usepackage{graphicx}
\usepackage{amsmath}
\usepackage{amsthm}
\usepackage{booktabs}
\usepackage{algorithm}
\usepackage{algorithmic}
\usepackage{xcolor}
\usepackage{tikz}
\usetikzlibrary{mindmap}
\usepackage{amssymb}
\usepackage{multirow}
\title{Augmenting Knowledge Graphs for Better Link Prediction}

\author{
Jiang Wang$^1$
\and
Filip Ilievski\and
Pedro Szekely\And
Ke-Thia Yao
\affiliations
USC Information Sciences Institute\\
\emails
\{jiangwan, ilievski, pszekely, kyao\}@isi.edu
}

\begin{document}

\maketitle

\begin{abstract}
Embedding methods have demonstrated robust performance on the task of link prediction in knowledge graphs, by mostly encoding entity relationships. Recent methods propose to enhance the loss function with a literal-aware term. In this paper, we propose \textsc{\textit{KGA}}: a knowledge graph augmentation method that incorporates literals in an embedding model without modifying its loss function. KGA discretizes quantity and year values into bins, and chains these bins both horizontally, modeling neighboring values, and vertically, modeling multiple levels of granularity. KGA is scalable and can be used as a pre-processing step for any existing knowledge graph embedding model. Experiments on legacy benchmarks and a new large benchmark, \textsc{\textit{DWD}}, show that augmenting the knowledge graph with quantities and years is beneficial for predicting both entities and numbers, as KGA outperforms the vanilla models and other relevant baselines. Our ablation studies confirm that both quantities and years contribute to KGA's performance, and that its performance depends on the discretization and binning settings. We make the code, models, and the DWD benchmark publicly available to facilitate reproducibility and future research.
\end{abstract}

\section{Introduction}
\label{sec:intro}


Hyperrelational knowledge graphs (KGs), like Wikidata~\cite{vrandevcic2014wikidata}, formalize knowledge as statements. A statement consists of a triple with key-value qualifiers and references that support its veracity. A statements represents either a relationship between two entities, or attributes a literal value (date, quantity, or string) to an entity. Nearly half of the statements in Wikidata are entity relationships, a third of them are string-valued, and the remaining (around 15\%) statements attribute quantities and dates to entities.\footnote{\url{https://tinyurl.com/56vyjd2y}, accessed on 13/01/22.} Intuitively, entity and literal statements complement each other to form a comprehensive view of an entity.


Considering the sparsity and the inherent incompleteness of KGs, the task of link prediction (LP) has been very popular, producing methods based on matrix factorization/decomposition, KG paths, and embeddings~\cite{wang2021survey}. While most LP methods only consider statements about two entities (Qnodes), some methods recognize the relevance of literals~\cite{gesese2019survey}. Literal-aware methods~\cite{lin2015learning,xiao2017ssp} predominantly learn a representation about the textual descriptions of an entity and combining it with its structured representation. 

Quantities and dates have seldom been considered, despite their critical role in contextualizing the LP task. A person's date of birth or a company's founding year anchor the prediction context to a specific time period, whereas the population of a country indicates how big a country is. Recent methods that incorporate numeric literals into KG embeddings add a literal-aware term to the scoring function~\cite{garcia2017kblrn,kristiadi2019incorporating}, or modify the loss function of the base model in order to balance capturing the KG structure and numeric literals~\cite{wu2018knowledge,tay2017multi,feng2019marine}. These methods leverage the knowledge encoded in literals in order to improve link prediction performance, but they introduce additional parameters and model-specific clauses, which limits their scalability and generalizability across embedding models.

In this paper, we investigate how to incorporate quantity and date literals into embedding-based link prediction models without modifying their loss function.\footnote{We focus on embedding-based methods, as these are superior over matrix factorization and path-based methods~\cite{lu2020utilizing}.} Our contributions are:

\noindent 1. \textsc{\textit{KGA}}: a \textbf{K}nowledge \textbf{G}raph \textbf{A}ugmentation method for incorporating quantity and year literals into KGs, by chaining the literals vertically, for different granularities of discretization, and horizontally, for neighboring values to each granularity level. KGA is scalable and can be used as a pre-processing step for any existing KG embedding model. 

\noindent 2. \textsc{\textit{DWD}}, an LP benchmark that is orders of magnitude larger than existing ones, and includes quantities and years, thus addressing evaluation challenges with size and overfitting. 

\noindent 3. An extensive LP evaluation, showing the superiority of KGA in terms of generalizability, scalability, and accuracy. Ablations study the individual impact of quantities and years, and the effect of discretization strategies and bin sizes.



\noindent 4. Public release of our code, resulting models, and the DWD benchmark: \url{https://github.com/Otamio/KGA/}.

\section{Task Definition}

We define \textit{knowledge graph with numeric triples} as a union of entity-valued and numeric statements, formally, $G = {(s,r,o)} \cup {(e,a,v)}$, where $(s,r,o) \in \mathcal{E} \times \mathcal{R} \times \mathcal{E} $ and ${(e,a,v)} \in \mathcal{E} \times \mathcal{R} \times \mathbb{R}$. The entity triples can be formalized as a set of triples (facts), each consisting of a relationship $r \in \mathcal{R}$ and two entities $s, o \in \mathcal{E}$, referred to as the \textit{subject} and \textit{object} of the triple. The numeric triples consist of an attribute $a \in \mathcal{R}$ with one entity $e \in \mathcal{E}$ and one value $v \in \mathbb{R}$.
We formalize \textit{entity link prediction} as a point-wise learning-to-rank task, with an objective to learn a scoring function $\psi : (\mathcal{E} \times \mathcal{R} \times \mathcal{E}) \rightarrow \mathbb{R}$. Given an input triple $x = (s, r, o)$, its score $\psi(x) \in \mathbb{R}$ is proportional to the likelihood that the fact encoded by $x$ is true.
We pose \textit{numeric link prediction} as a point-wise prediction task, with an objective to learn a function $\phi : (\mathcal{E} \times \mathcal{R}) \rightarrow \mathbb{R}$. Given an entity-attribute pair, $(e, a)$, the output $\phi (e,a)$ is the numeric value of attribute $a$ of entity $e$.



\begin{figure}
    \centering
    \includegraphics[width=\linewidth]{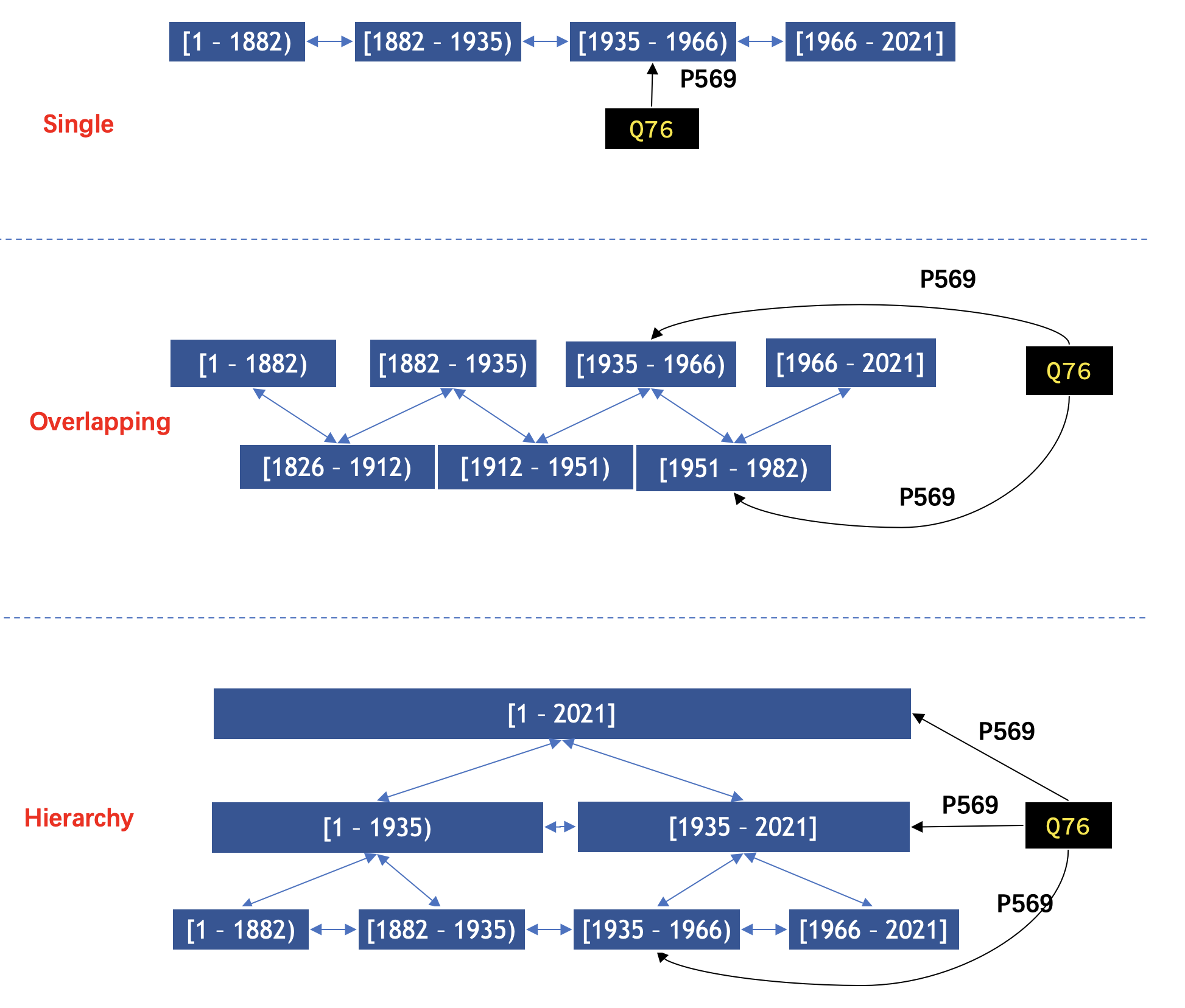}
    \caption{Single (top), overlapping (middle), and hierarchy (bottom) binning for the attribute triple (Q76, P569, ``1961''), which specifies the year of birth of Barack Obama. We use $b=4$ bins, quantile-based binning, and chaining in each level mode.} 
    \label{fig:example}
\end{figure}

\section{KGA}

The key idea of KGA is to augment the KG with quantities and years before training. KGA consists of two steps: 1) discretization and bin creation; and 2) graph augmentation. 

The goal of the first step is to discretize the entire space of values for a KG attribute into bins. Binning significantly decreases the number of distinct values that a model needs to learn, thus reducing the representational and computational complexity, and improving the model performance for relationships with sparse data.
For instance, the entire set of birth date (P569) values in Wikidata could be split based on value frequency into four bins: $[1 - 1882), [1882 - 1935), [1935 - 1966),$ and $[1966 - 2021]$ (Figure~\ref{fig:example}, top).\footnote{All year values in Wikidata are positive. Years BC are also positive, but include an additional qualifier to indicate the era.} 
In total, KGA defines two different bin interval settings, based on either interval width (fixed) or value frequency (quantile), and three methods to set the bin levels: single, overlapping, and hierarchy.
In this example, we use frequency to set the bin intervals and we use a single series of bins. 
While discretizing numeric values into bins is not new, KGA is the first method that uses binning of numeric values to improve LP performance. 

In the augmentation step, each original value is assigned to its bin, and its binned value is used to augment the KG. For example, Barack Obama's year of birth, 1961, belongs to the third bin ($[1935 - 1966)$), so the original triple (Q76, P569, ``1961'') will be translated into (Q76, P569, $[1935 - 1966)$). Besides the assignment of the correct bin to its entity, KGA also allows for the neighboring bins to be chained to each other. The augmented KG, containing the original entity relationships and the added binned literal knowledge, can then be used to train any standard off-the-shelf embedding model. 

We detail KGA's two-step approach, and its use for LP. 


\subsection{Discretization and Bin Creation}

We start by obtaining the set of values $v_i$ which are associated with entities $e\in {E}$ of a given attribute $a$, namely $(e, a, v_i) \in G$. As dates come in different granularities (e.g., year, month, or day), we transform all dates with a granularity of year or finer (e.g., month) to year values. We sort these values in an ascending order, obtaining $V=\{v_1, v_2, ..., v_m\}$, such that $v_i\leq v_{i+1}$. We next describe how we create bins based on the values in $V$, resulting in a dictionary $D: V \rightarrow \mathcal{E}$ per attribute, which maps each value in $V$ to one or multiple bin entities in $\mathcal{E}$.
The discretization of KGA consists of two components: 1) definition of bin intervals; and 2) specification of bin levels.





\noindent \textbf{1. Bin intervals} specify the start and end values for the bins of a given attribute. Formally, given a minimal attribute value of $z^-$, a maximal value of $z^+$, and a set of $b$ bins, we seek to define $[z_i^-,z_i^+)$, for each bin $i$, where $i\in[0,b)$. Here, $b$ is a predefined parameter. We investigate two bin interval settings: fixed intervals and quantile-based intervals.\footnote{We also experimented with methods that group values based on their (cross-)group (dis)similarity like k-means clustering~\cite{lloyd82}, but we do not report these given their low performance.}

Our \textit{fixed} (equal width) binning strategy divides the entire space of values into bins of equal width. The length of each bin $i$ is $k=\frac{(z^+-z^-)}{b}$, and the $i$-th bin contains the values $z_i=[z^-+ik,z^-+(i+1)k)$.

The \textit{quantile} (equal frequency) binning strategy splits [$z^-,z^+$] based on the distributional density, by ensuring that each bin $i$ covers the values for the same number of entities. This allows denser areas of the distribution to be represented with more bins. For a attribute $a$ defined for $N_a$ entities, each bin will cover approximately $n=\frac{N_a}{b}$ entities. 







\noindent \textbf{2. Bin levels} specify whether the binning method will create a single set of bins, two overlapping sets of bins, or a hierarchy with several levels of bins.

The \textit{single} strategy creates only one set of disjoint bins, as described previously. 

The \textit{overlapping} strategy creates bins with intervals that overlap. Starting from a single auxiliary set of $2b$ bins, the overlapping bins are created by merging two neighboring bins at a time. The $i$-th merged bin consists of merging bin $b_i$ and $b_{i+1}$, the $(i+1)$-th merges $b_{i+1}$ and $b_{i+2}$, and so on. This process results in $2b-1$ overlapping bins. Overlapping bins are illustrated in Figure~\ref{fig:example}, middle. Here, the auxiliary bins (not shown) would be $[[1, 1826], [1826, 1882], [1882, 1912], …]$. The derived overlapping bins merge two neighboring bins at a time - merging the first two bins produces $[1,1882]$, merging the second and the third produces $[1826, 1912]$, etc.
%

The \textit{hierarchy} strategy creates multiple binning levels, signifying different granularity levels. A level $l$ has $b^l$ bins. The $0$-th level is the coarsest, and it contains a single bin with the entire set of values $[z^-,z^+)$. The levels grow further in terms of detail: considering $b=2$, the first level has 2 bins, the second one has 4, and the third one has 8. An example for a quantile-based hierarchy binning is shown in Figure~\ref{fig:example} bottom.
The entire set of birth years at level $0$ ($[1, 2021]$) is split into two bins at level $1$: $[1, 1935)$ and $[1935, 2021]$. Level $2$ splits the bins at level $1$ into even smaller, more precise bins: $[1,1882)$, $[1182, 1935)$, $[1935, 1966)$ and $[1966, 2021]$.

\subsection{Graph Augmentation}

The created bins provide a mapping from a value $v_i \in V$ to a set of bins $b_i \in B$. In order to augment the original graph $G$, we perform two operations: chaining of bins and linking bins to corresponding entities.

\textbf{1. Bin chaining} defines connections between two bins $b_i$ and $b_j$, where $b_i, b_j \in B$. The links between bins depend on the bin level setting. Both single bins and overlapping bins are chained by connecting each bin $b_i$ to its predecessor $b_{i-1}$ and successor $b_{i+1}$, see Figure~\ref{fig:example} (top and middle).
The bins in the hierarchy augmentation mode are connected both horizontally and vertically: 1) horizontally each bin $b_i$ connects to its predecessor $b_{i-1}$ and successor $b_{i+1}$; 2) vertically each bin is connected to its coarser level and finer level correspondents (see Figure~\ref{fig:example} bottom).



\textbf{2. Bin assignment} links the generated bins $B(v)$ for a given attribute value $v$ to its corresponding entity $e$. A numeric triple $(e, a, v)$ is replaced with a set of triples $(e, a, b)$, where $b \in B(v)$. The updated set of triples modifies the original graph $G$ into an augmented graph $G^\prime$.
Figure~\ref{fig:example} presents examples of bin assignment. The birth year (attribute P569) of Barack Obama (entity Q76) is $1961$. 
For single level bins (Figure~\ref{fig:example}, top), the birth year of Obama is placed into bin $[1935, 1966)$. For overlapping bins (Figure~\ref{fig:example}, middle), the birth year is placed into the bins $[1935, 1966)$ and $[1951, 1982)$. And, when using hierarchy bin levels (Figure~\ref{fig:example}, bottom), the birth year is placed into bins $[1, 2021]$, $[1935, 2021]$ and $[1935, 1966)$.

\subsection{Link Prediction} $G\prime$ can now be used to perform entity and numeric LP. KGA is trained by simply replacing $G$ with $G^\prime$ in the input to the base embedding model. Link prediction of entities with KGA is performed by selecting the entity node $e$ with a highest probability. Numeric link prediction is performed by selecting the bin $b$ with the highest score as a predicted range; and obtaining its median, $m$, as an approximate predicted value.



\section{Experimental setup}

\subsection{Datasets and Evaluation}

We use benchmarks based on three KGs to evaluate LP performance. We show data statistics in Table~\ref{tab:years_yago} in the appendix.

1. \textbf{FB15K-237}~\cite{toutanova2015observed} 
is a subset of Freebase~\cite{bollacker2008freebase}, which mainly covers facts about movies, actors, awards, sports, and sports teams. FB15K-237 is derived from the benchmark FB15K by removing inverse relations, because a significant number of test triples in FB15K could be inferred by simply reversing the triples in the training set~\cite{dettmers2018convolutional}. We use the same split as \cite{toutanova2015observed}. 
For numeric prediction we follow \cite{kotnis2019learning}. 


2. \textbf{YAGO15K}~\cite{liu2019mmkg} 
is a subset of YAGO~\cite{suchanek2007yago}, which is also a general-domain KG. YAGO15K contains 40\% of the data in the FB15K-237 dataset. However, YAGO15K contains more valid numeric triples than FB15K-237. For entity LP, we use the data split proposed in~\cite{lacroix2020tensor}, while for numeric prediction we follow \cite{kotnis2019learning}. 

3. \textbf{DWD}. We introduce DARPA Wikidata (DWD), a large subset of Wikidata~\cite{vrandevcic2014wikidata} that excludes prominent domain-specific Wikidata classes such as review article (Q7318358), scholarly Article (Q13442814), and chemical compound (Q11173). DWD describes over 37M items (42\% of all items in Wikidata) with approximately 166M statements. DWD is several orders of magnitude larger than any of the previous LP benchmarks, thus providing a realistic evaluation dataset with sparsity and size akin to the size of modern hyperrelational KGs. 
We split the DWD benchmark at a 98-1-1 ratio, for both entity and numeric link prediction. Given the size of the DWD benchmark, 1\% corresponds to a large number of data points (over 1M statements).



For entity LP on FB15K-237 and YAGO15K, we preserve a checkpoint of the model every 5 epochs for DistMult, ComplEx, every 10 epochs for ConvE, TuckER, and every 50 steps for TransE, RotatE. We use the MRR on the validation set to select the best model checkpoint. We report the filtered MRR, Hits@1, and Hits@10 of the best model checkpoint. For DWD, we preserve a checkpoint for each epoch. We use the MRR on the validation set to select the best model checkpoint, and report unfiltered MRR and Hits@10~\cite{MLSYS2019_e2c420d9}. Filtered MRR evaluation requires discarding all positive edges when generating corrupted triples, which is not scalable for large KGs. 
For DWD, we report these metrics by ranking positive edges among $C=500$ randomly sampled corrupted edges. 
For numeric LP, we report MAE for each KG attribute. 

\subsection{Models}

For entity link prediction, we evaluate \textit{KGA} with the following six embedding models: TransE~\cite{bordes2013translating},
DistMult~\cite{yang2014embedding}, ComplEx~\cite{trouillon2016complex}, ConvE~\cite{dettmers2018convolutional}, RotatE~\cite{sun2019rotate}, and TuckER~\cite{balavzevic2019tucker}. 
We run KGA with $b \in [2,4,8,16,32]$ and show the best result for each model.
We compare KGA to the embedding-based LP predictions on the original graph, and to prior literal-aware methods: KBLN~\cite{garcia2017kblrn}, MTKGNN~\cite{tay2017multi}, and LiteralE~\cite{kristiadi2019incorporating}. 
We compare our numeric LP performance to the methods NAP++ ~\cite{kotnis2019learning} and MrAP \cite{bayram2021node}. As NAP++ is based on TransE embeddings, we focus on our results with this embedding model, and we provide the results for the remaining five embedding models in the appendix. As mentioned above, we take the median of the most probable bin to be the predicted value by KGA.

The more computationally demanding embedding models (ConvE, RotatE, and TuckER) cannot be run on DWD. The size of DWD is prohibitive for ConvE and TuckER because they depend on 1-N sampling, where batch training requires to load the entire entity embedding into memory.\footnote{With a hidden size of 200, loading the entity embedding in memory requires at least 42.5M * 200 * 4 = 34 GB GPU memory, which is challenging for most GPU devices today.} RotatE is even more memory-intensive, because of its hidden size of 2000, which requires an order of magnitude more memory compared to ConvE. As LiteralE and KBLN are based on these methods, they can also not run on DWD (see appendix for more implementation details).
Thus, on DWD, we compare KGA's performance on the models ComplEx, TransE, and DistMult against their base model performance.




We provide details on the parameter values, software, and computational environments in the technical appendix.


\section{Results}

\subsection{Entity Link Prediction Results}

\textbf{Main results} LP results on FB15K-237 and YAGO15K are shown in Table~\ref{tab:link_prediction}.  KGA outperforms the vanilla model and the literal-aware LP methods by a comfortable margin with all six embedding models. 
The best overall performance is obtained with the TuckER and RotatE models for FB15K-237, and the ConvE and RotatE models for YAGO15K, which is in line with the relative results of the original models. On FB15K237, the largest performance gain is obtained for DistMult (2.7 MRR points), and the performance gain is around 1 MRR point for the remaining models. The comparatively smaller improvement for TuckER on FB15K-237, brought by KGA and other literal-aware methods, could be due to TuckER's model parameters being particularly tuned for this dataset. This intuition is supported by the much higher contribution of KGA for TuckER on the YAGO15K dataset. KGA brings a higher MRR gain on the YAGO15K dataset, which shows that KGA is able to learn valuable information from the additional 24\% literal triples which we added to YAGO15K. On YAGO15K, KGA improves the performance of the most recent models (ConvE, RotatE, and TuckER) by over 2 MRR points. These results highlight the impact of KGA's approach of augmenting KG embedding models with discretized literal values.

\begin{table}[!t]
    \centering
    \small
    \caption{LP results on FB15K-237 and YAGO15K. We compare KGA to the original model (-), and the baselines LiteralE and KBLN. We report the reproduced results for all baseline methods, and provide the original results in the appendix. For KGA, we show the best results across discretization strategies (single, overlapping, hierarchy) and numbers of bins (2, 4, 8, 16, 32). We bold the best overall result per metric, and underline the best result per model.}
    \label{tab:link_prediction}
    {\footnotesize
    \begin{tabular}{l | l l l | r r r }
     & \multicolumn{3}{c|}{\bf FB15K-237} & \multicolumn{3}{c}{\bf YAGO15K} \\ 
    \bf method & MRR & H@1 & H@10 & MRR & H@1 & H@10 \\ 
    \hline
    \bf TransE      &   0.315   &   0.217   &   0.508   &   0.459   &   0.376   &   0.615   \\
    +LiteralE       &   0.315   &   0.218   &   0.504   &   0.458   &   0.376   &   0.612   \\
    +KBLN           &   0.308   &   0.210   &   0.496   &   0.466   &   0.382   &   0.621   \\
    +KGA            &   \underline{0.321}   &   \underline{0.223}   &   \underline{0.516}   &   \underline{0.470}   &   \underline{0.387}   &   \underline{0.623}   \\
    \hline
    \bf DistMult    &   0.295   &   0.212   &   0.463   &   0.457   &   0.389   &   0.585   \\
    +LiteralE       &   0.309   &   0.223   &   0.481   &   0.462   &   0.396   &   0.587   \\
    +KBLN           &   0.302   &   0.220   &   0.470   &   0.449   &   0.377   &   0.581   \\
    +KGA            &   \underline{0.322}   &   \underline{0.233}   &   \underline{0.502}   &   \underline{0.472}   &   \underline{0.402}   &   \underline{0.606}   \\
    \hline
    \bf ComplEx     &   0.288   &   0.205   &   0.455   &   0.441   &   0.370   &   0.572   \\
    +LiteralE       &   0.295   &   0.212   &   0.462   &   0.443   &   0.375   &   0.570   \\
    +KBLN           &   0.293   &   0.213   &   0.451   &   0.451   &   0.380   &   0.583   \\
    +KGA            &   \underline{0.305}   &   \underline{0.219} &   \underline{0.478}   &   \underline{0.453}   &   \underline{0.380}   &   \underline{0.591}   \\
    \hline
    \bf ConvE       &   0.314   &   0.226   &   0.488   &   0.470   &   0.405   &   0.597   \\
    +LiteralE       &   0.317   &   0.230   &   0.489   &   0.475   &   0.408   &   0.601   \\
    +KBLN           &   0.305   &   0.219   &   0.479   &   0.474   &   0.408   &   0.600   \\
    +KGA            &   \underline{0.329}   &   \underline{0.239}   &   \underline{0.512}   &   \bf \underline{0.492}   &   \bf \underline{0.427}   &   \underline{0.616}   \\
    \hline
    \bf RotatE      &   0.324   &   0.232   &   0.506   &   0.451   &   0.370   &   0.605   \\
    +LiteralE       &   0.329   &   0.237   &   0.512   &   \underline{0.475}   &   \underline{0.400}   &   0.619   \\
    +KBLN           &   0.314   &   0.222   &   0.500   &   0.469   &   0.393   &   0.613   \\
    +KGA            &   \underline{0.335}   &   \underline{0.242}   &   \underline{0.521}   &   0.473   &   0.392   &   \bf \underline{0.626}   \\ 
    \hline
    \bf TuckER      &   0.354   &   0.263   &   0.536   &   0.433   &   0.360   &   0.571   \\
    +LiteralE       &   0.353   &   0.262   &   0.536   &   0.421   &   0.348   &   0.564   \\
    +KBLN           &   0.345   &   0.253   &   0.530   &   0.420   &   0.349   &   0.556   \\
    +KGA            &   \bf \underline{0.357}   &   \bf \underline{0.265}   &   \bf \underline{0.540}   &   \underline{0.454}   &   \underline{0.380}   &  \underline{0.592}   \\ 
    \hline
    \end{tabular}
    }

\end{table}

\begin{table}[!t]
    \centering
    \small
    \caption{LP results on DWD. We show the performance (MRR and Hits@10) of the vanilla embedding model (-), and KGA with binned quantities, with years, and the full KGA. We use 32-bin KGA with QOC (quantile, overlapping, and chaining) discretization.} 
    \label{tab:lp_dwd}
    \begin{tabular}{l | r r | r r | r r}
    \bf Method & \multicolumn{2}{c}{\bf TransE} & \multicolumn{2}{c}{\bf DistMult} & \multicolumn{2}{c}{\bf ComplEx} \\ 
     & MRR & H@10 & MRR & H@10 & MRR & H@10 \\ 
    \hline
      -         &   0.580   &   0.762   &   0.559   &   0.740   &   0.568   &   0.746   \\
      Quantity  &   0.582   &   0.764   &   0.564   &   0.744   &   0.571   &   0.748   \\ 
      Year      &   0.580   &   0.763   &   0.562   &   0.744   &   0.569   &   0.747   \\
      KGA       &   \bf 0.583   &   \bf 0.764   &   \bf 0.566   &   \bf 0.746   &   \bf 0.574   &   \bf 0.751   \\ 
      \hline
    \end{tabular}
\end{table}

\noindent \textbf{Scalability and knowledge ablations} We test the ability of KGA to perform LP on DWD. The results (Table~\ref{tab:lp_dwd}) reveal that KGA, unlike prior literal-aware methods, scales well to much larger LP benchmarks. Furthermore, we observe that KGA brings steady improvement across the models and the metrics. Our knowledge ablation study shows that integrating either quantities or years is better than the vanilla model, that quantities are marginally more informative than years, and that integrating both of them yields best performance. We conclude that quantities and years are informative and mutually complementary for LP by embedding models. 

\begin{table*}[!t]
    \centering
    \small
    \caption{Ablation study on modes of graph augmentation with link prediction on FB15K-237. KGA variants: `-' represents the original graph (no augmentation), F = Fixed Size, Q = Quantile, S = Single, O = Overlapping, H = Hierarchy, C = Chaining,  N = No Chaining. The best result for each column is marked in bold. We show the best results among the different numbers of bins (2, 4, 8, 16, 32).}
    \label{tab:ablation_modes}
    \begin{tabular}{l | r r | r r | r r | r r | r r | r r }
      & \multicolumn{2}{c}{\bf TransE} & \multicolumn{2}{c}{\bf DistMult} & \multicolumn{2}{c}{\bf ComplEx} & \multicolumn{2}{c}{\bf ConvE} & \multicolumn{2}{c}{\bf RotatE} & \multicolumn{2}{c}{\bf TuckER}  \\ 
      \bf KGA       &   MRR     &   H@10    &   MRR     &   H@10    &   MRR     &   H@10     & MRR    & H@10  & MRR    & H@10  & MRR    & H@10   \\
      \hline
      \bf -         &   0.315   &   0.508   &   0.295   &   0.463   &   0.288   &   0.455 & 0.314  & 0.488 & 0.324  & 0.506 & 0.354  & 0.536   \\
      \hline
      \bf FSC       &   0.317   &   0.509   &   0.301   &   0.471   &   0.291   &   0.459 & 0.320  & 0.494 & 0.328  & 0.513 & 0.354  & 0.536   \\
      \bf FOC       &   0.318   &   0.512   &   0.306   &   0.482   &   0.296   &   0.466  & 0.319  & 0.494 & 0.327  & 0.511 & 0.354  & 0.536 \\
      \bf FHC       &   0.318   &   0.511   &   0.304   &   0.478   &   0.299   &   0.469  & 0.320  & 0.495 & 0.327  & 0.510 & 0.354  & 0.535  \\
      \bf FON       &   0.319   &   0.510   &   0.305   &   0.480   &   0.296   &   0.464  & 0.321  & 0.498 & 0.328  & 0.513 & 0.353  & 0.535   \\
      \hline 
      \bf QSC       &   0.320   &   0.513   &   0.303   &   0.475   &   0.296   &   0.465  & 0.319  & 0.494 & 0.330  &   0.516   & \bf 0.357  & 0.540   \\
      \bf QOC       &   0.321   &   0.513   &   0.312   &   0.487   &   0.299   &   0.471  & 0.322  & 0.499 & 0.332  & 0.517 & 0.356  & \bf 0.542 \\
      \bf QHC       & \bf 0.321   & \bf  0.516   &   \bf 0.322   &   \bf 0.502  &  \bf  0.305   & \bf 0.478  & \bf 0.329  & \bf 0.512 & \bf 0.335  & \bf 0.521 & 0.356  & 0.538   \\
      \bf QON       &   0.320   &   0.514   &   0.309   &   0.480   &   0.299   &   0.468  & 0.321  & 0.498 & 0.332  & 0.516 & 0.355  & 0.536  \\
      \hline
    \end{tabular}

\end{table*}

\begin{table}[!t]
    \centering
    \small
    \caption{Effect of bin size on the performance of different models on FB15K-237. We show results for the best discretization strategy. We experiment with 2, 4, 8, 16, and 32 bins. Numbers indicate MRR. }
    \label{tab:bin_analysis}
    \begin{tabular}{l | r r r r r}
    \bf model       & \bf 2 & \bf 4  & \bf 8  & \bf 16  & \bf 32  \\
    \hline
    TransE          &   0.321   &   0.320   &   0.321   &   \bf 0.321   &   0.321   \\
    DistMult        &   0.306   &   0.308   &   0.314   &   0.317   &   \bf 0.322   \\
    ComplEx         &   0.294   &   0.295   &   0.300   &   0.304   &   \bf 0.305   \\
    ConvE           &   0.321   &   0.320   &   0.325   &   0.325   &   \bf 0.329   \\
    RotatE          &   0.327   &   0.326   &   0.332   &   \bf 0.335   &   0.334   \\
    TuckER          &   0.354   &   0.356   &   0.355   &   0.356   &   \bf 0.357   \\
    \hline
    \end{tabular}
\end{table}

However, the improvement brought by KGA for the models TransE, DistMult, and ComplEx is between 0.3 and 0.7 MRR points, which is much lower than its impact on the smaller datasets. We note that the results in Table~\ref{tab:link_prediction} show the best configuration for KGA, while the results in Table~\ref{tab:lp_dwd} show a single configuration (QOC with 32 bins). Therefore, we hypothesize that the performance of KGA on DWD can be improved with further tuning of the number of bins and the discretization strategies applied. For computational reasons, we investigate the impact bin sizes and discretization strategies on the FB15K dataset, and leave the analogous investigation for DWD to future work.

\noindent \textbf{Binning ablations} 
We study different variants of discretization (Fixed and Quantile-based), bin levels (Single, Overlapping, and Hierarchy), and bin sizes (2, 4, 8, 16, and 32) on the FB15K-237 benchmark. The results (Table~\ref{tab:ablation_modes}) show that all of the discretization variants of KGA consistently improve over the baseline model. We observe that quantile-based binning is superior over fixed width binning, which is intuitive because quantile binning considers the density of the value distribution. Among the bin levels, we see that hierarchy bin levels performs better than overlapping binning, which in turn performs better than using a single bin. This relative order of performance correlates with the expressivity of each bin levels strategy. Comparing the quantile-overlapping versions with and without chaining, we typically observe a small benefit of chaining the bins horizontally. Table~\ref{tab:bin_analysis} provides results of KGA with different bin sizes (2, 4, 8, 16, 32) for all six models. We observe that finer-grained binning is generally preferred, as 32 bins performs best for 4 of the 6 models. Yet, we observe that for RotatE, the best performance is obtained with 16 bins, whereas for TransE, the number of bins has no measurable impact.
We conclude that the performance of KGA depends on selecting the best discretization strategy and bin size. While more expressive discretization and fine-grained binning generally works better, the optimal configuration is model-dependent and should be investigated further. 

\subsection{Numeric Link Prediction Results}

\begin{table}[!t]
    \centering
    \small
    \caption{Performance of our numeric predictor with different choices of base model on graph augmented with 32-bin QOC, when compared to existing SOTA methods on the FB15K-237 and YAGO15K dataset. Numbers indicate MAE. 
    Values of NAP++ and MrAP are taken from~\protect\cite{bayram2021node}. We show results for KGA with TransE for a fair comparison to NAP++.} 
    \label{tab:value_imputation}
    \begin{tabular}{l | l r r r }
    & & \bf KGA & \bf NAP++ & \bf MrAP \\
     \hline
     \multirow{11}{*}{FB15K-237} 
     & date\_of\_birth       & 18.9         &   22.1      & \bf 15.0    \\
     & date\_of\_death       & 20.6         &   52.3      & \bf 16.3    \\
     & film\_release         & \bf 4.0      &    9.9      & 6.3         \\
     & organization\_founded & \bf 49.0     &   59.3      & 58.3        \\
     & location\_founded     & \bf 76.0     &   92.1      & 98.8        \\
     & latitude              & 2.1          &   11.8      & \bf 1.5     \\
     & longitude             & 7.1          &   54.7      & \bf 4.0     \\
     & area                  & \bf 6.1e4    &   4.4e5     & 4.4e5       \\
     & population            & \bf 4.0e6    &   7.5e6     & 2.1e7       \\
     & height                & \bf 0.077    &   0.080     & 0.086       \\
     & weight                & \bf 11.6     &   15.3      & 12.9        \\ 
     \hline
     \multirow{7}{*}{YAGO15K} 
     & date\_of\_birth       & \bf 16.3     & 23.2      & 19.7      \\
     & date\_of\_death       & \bf 30.8     & 45.7      & 34.0      \\
     & date\_created         & \bf 58.2     & 83.5      & 70.4      \\
     & data\_destroyed       & \bf 23.3     & 38.2      & 34.6      \\
     & date\_happened        & \bf 29.9     & 73.7      & 54.1      \\
     & latitude              & 3.4          & 8.7       & \bf 2.8   \\
     & longitude             & 7.2          & 43.1      & \bf 5.7   \\ 
     \hline
    \end{tabular}
\end{table}

\textbf{Main results} Next, we investigate the ability of KGA to predict quantity and year values directly. We compare KGA-QOC with 32 bins to the baselines MrAP and NAP++ on the FB15K-237 and YAGO15K benchmarks in Table~\ref{tab:value_imputation}. We observe that both KGA and MrAP perform better than NAP++.\footnote{The results from the original NAP++ paper differ from those obtained by reproducing experiments, see~\cite{bayram2021node}.} KGA performs better than MrAP on the majority of the attributes: 7 out of 11 in FB15K-237, and 5 out of 7 attributes in YAGO15K. Closer investigation reveals that our method is superior in decreasing the error for years and attributes with a large range of values, such as area and population, which could be attributed to the quantile-based binning strategy. MrAP outperforms our model on the latitude and longitude attributes on both datasets. Given the low MAE error of MrAP, we think that such a regression model that operates on the raw numeric values of triples can learn to reliably predict latitude and longitudes based on other structured information captured by the base embedding model, while literal information might. notimprove prediction further. 

We show the numeric LP results for the five most populous quantity attributes and the five most populous year attributes in DWD in Table~\ref{tab:dwd_quantities}. We compare KGA and a linear regression (LR) model to a baseline which selects the median value for an attribute. Both KGA and the LR model perform better than the median baseline, yet, the LR model outperforms KGA on this dataset for 9 out of 10 attributes. These results reinforce the findings in Table~\ref{tab:lp_dwd} that KGA requires further tuning in order to be applied to DWD.

\begin{table}[!t]
    \centering
    \small
    \caption{Performance of our numeric predictor KGA-QOC on DWD compared to a linear regression (LR) model and a median baseline. We use 32 bins for both quantities and years. Numbers indicate MAE reduction percentages against a median value baseline. We report results for the most populous 5 properties for both quantities and years, with identifiers: P1087, P6258$|$Q28390, P2044$|$Q11573, P6257$|$Q28390, P1215, P569, P570, P577, P571, and P585.}
    \label{tab:dwd_quantities}
    \begin{tabular}{l | r r r }
    \bf attribute               & \bf Median  & \bf LR    & \bf KGA  \\
    \hline
    Elo rating                  &   119.03      &   86.09       &   \bf 55.20   \\
    declination (degree)        &   18.68       &   \bf 9.83        &   18.53   \\
    elevation above sea level   &   466.51      &   \bf 366.64      &   459.48  \\
    right ascension (degree)    &   82.98       &   \bf 40.90       &   82.51   \\
    apparent magnitude          &   3.02        &   \bf 2.00        &   2.37    \\
    \hline
    date of birth               &  62.71        & \bf 49.70         &   58.59   \\
    date of death               &  90.68        & \bf 78.10         &   79.38   \\
    publication date            &  28.33        & \bf 17.37         &   28.27   \\
    inception                   &  72.84        & \bf 61.45         &   72.27   \\
    point in time               &  88.76        & \bf 81.65         &   83.70   \\
    \hline
    \end{tabular}
\end{table}

\section{Related work}


\noindent \textbf{Graph Embedding with Literals}
Literal-aware LP methods~\cite{lin2015learning,xiao2017ssp} predominantly focus on strings, by learning a representation of the textual descriptions of an entity and combining it with its structured representation~\cite{gesese2019survey}. Considering string-valued triples is a natural future extension of KGA. Several efforts incorporate numeric triples into KG embeddings by adding a literal-aware term to the scoring function of the embedding model. LiteralE~\cite{kristiadi2019incorporating} incorporates literals by passing literal-enriched embeddings to the scoring function. 
Assuming that the difference between the numeric values for a relation is a good indicator of the existence of a relation, KBLN~\cite{garcia2017kblrn} adds a separate scoring function for literals.
Our experiments show that KGA performs better than both LiteralE and KBLN on entity LP.
Instead of modifying the scoring function, several methods modify the loss function of the base model to balance between predicting numeric and entity values.
TransEA~\cite{wu2018knowledge} extends TransE with a regression penalty on the base model, while MTKGNN~\cite{tay2017multi} uses multitask learning and extends a neural representation learning baseline by introducing separate training steps that use embedding to predict numeric values. MARINE~\cite{feng2019marine} extends these methods by adding a proximity loss, which preserves the embedding similarity based on shared neighbors between two nodes. We do not compare against TransEA and MARINE because their reported performance is lower than recent base models or KBLN, whereas comparison to MTKGNN would require re-implementation of the model, as the original work is evaluated on different datasets and has its own code base. In contrast to prior work, KGA augments the structure of the original KG, leaving the loss function of the base model intact. As a consequence, KGA can be directly reused to new embedding methods without customizing the base algorithm or the scoring function, and it can be computed on large KGs with the size of Wikidata. Furthermore, the explicitly represented literal range values are intrinsically meaningful as intuitive approximation, corresponding to how humans perceive numbers~\cite{dehaene2011number}.



\noindent \textbf{Numeric Link Prediction}
MTKGNN~\cite{tay2017multi} proposes a multitask learning algorithm that predicts statements with entity and numeric values. NAP++~\cite{kotnis2019learning} uses TransEA~\cite{wu2018knowledge} to cluster embeddings based on numeric triples and a relation propagation algorithm to predict attribute values. MrAP~\cite{bayram2021node} uses message passing within and between entities to predict the attributes with a strong normal distribution. 
KGA also predicts numeric values, by discretizing them into buckets, using the base algorithm to predict the correct bin, and selecting the median value of the predicted bin. Several other work has also considered numeric LP. KGA is more controllable and intuitive, allowing its users to understand what the embedding has learned through simple LP. 

\section{Conclusions}

This paper proposed a knowledge graph augmentation (KGA) method, which incorporates literals into embedding-based link prediction systems in a pre-processing step. KGA does not modify the scoring or the loss function of the model, instead, it enhances the original KG with discretized quantities and years. Thus, KGA is designed to generalize to any embedding model and KG. We formulated variants of KGA that differ in terms of their interval definition, binning levels, number of bins, and link formalization. Evaluation showed the superiority of KGA over vanilla embedding models and baseline methods on both entity and numeric link prediction. Unlike prior baselines, KGA scaled to a Wikidata-sized KG, as it performs a minor adaptation of the original model. The performance of KGA depends on the selected number of bins and discretization strategy. While more expressive discretization and binning usually fares better, optimal performance is model-dependent and should be investigated further. Future work should also extend KGA to include string literals, and to enhance link prediction on other graphs, like DBpedia.

\bibliographystyle{named}
\bibliography{ijcai22}

\newpage
\clearpage
\appendix
\section*{Appendix}

\subsection*{Data statistics}

\begin{table}[!t]
    \small
    \caption{Statistics of the benchmarks used in this paper.} 
    \label{tab:years_yago}
    \begin{tabular}{l | r r r}
     \bf dataset        & \bf FB15K-237         & \bf YAGO15K & \bf DWD         \\ \hline
      \# Entities       & 14,541                &  15,136     & 42,575,933      \\
      \# Relations      & 237                   &  32         & 1,335           \\
      \# Triples        & 310,116               &  98,308     & 182,246,241     \\
      \# Attributes     & 116                   &  7          & 565             \\
      \# Literals       & 29,220                &  23,520     & 31,925,813      \\ \hline
    \end{tabular}

\end{table}

We provide statistics for our evaluation datasets in Table~\ref{tab:years_yago}.

\subsection*{Implementation Details}


For FB15K237 and YAGO15k, we run our experiments on top of the codebase of \cite{sun2019rotate} (TransE, RotatE), and \cite{balavzevic2019tucker} (DistMult, ComplEx, ConvE, TuckER). We extend the six base models and implement LiteralE and KBLN for both codebases. Since the training and sampling scheme of LiteralE and TuckER are highly similar, we introduced the models from LiteralE to the TuckER codebase with little change ~\cite{kristiadi2019incorporating}. For the RotatE codebase, for LiteralE, the literal-enriched embedding is computed per epoch to make it computation feasible, and for KBLN, we only modified the scoring function so that it is directly comparable with the TuckER implementation. We have shared the code publicly for readers' evaluation.

Because memory and computation constraints, we are not able to evaluate the performance of large datasets with the RotatE and the TuckER codebase. For the RotatE codebase, memory will explode when the code calls the evaluator; for the TuckER codebase, memory will explode with the large matrix training per batch, since it uses 1-N sampling scheme. To address the above issues, we use Pytorch-Biggraph~\cite{MLSYS2019_e2c420d9} to run TransE, DistMult, and ComplEx on the DWD dataset. We also use Pytorch-BigGraph to create the embedding that will be used in numeric link prediction. While we are not able to offer a direct comparison between LiteralE, KBLN, and KGA, we have offered readers a comparison between KGA and vanilla embedding for evaluation. 

For experiments on FB15K-237 and YAGO15K, we run the experiments on a 32-core, 256 GB RAM server with a Quadro RTX 8,000 GPU. For the experiments on DWD, we conduct the experiments on a 64-core, 1TB RAM server with only CPUs, installed with the newest version of PyTorch-BigGraph.

\subsection*{Parameter values}

For experiments on FB15K237 and YAGO15K, we used the following parameters:

1. \textbf{TransE}. Batch size of 1024, number of steps of 1,500, learning rate of 0.0001, number of negatives of 256, embedding dimension of 1000, gamma of 24, and adversial temperature of 1.0.

2. \textbf{DistMult}. Batch size of 128, number of epochs of 200, learning rate of 0.003, decay rate of 0.995, embedding dimension of 200, input dropout of 0.2, and label smoothing of 0.1. 

3. \textbf{ComplEx}. Batch size of 128, number of epochs of 200, learning rate of 0.003, decay rate of 0.995, embedding dimension of 400 (200 for real and 200 for imaginary), input dropout of 0.2, and label smoothing of 0.1. 

4. \textbf{ConvE}. Batch size of 128, number of epochs of 1000, learning rate of 0.003, decay rate of 0.995, embedding dimension of 200, input dropout of 0.2, hidden dropout of 0.3, an feature map dropout of 0.2, and label smoothing of 0.1. For other parameters, we just use the same values listed by \cite{dettmers2018convolutional}.

5. \textbf{RotatE}. Batch size of 1024, number of steps of 1,500, learning rate of 0.0001, number of negatives of 256, embedding dimension of 1000, gamma of 24, and adversial temperature of 1.0.

6. \textbf{TuckER}. For FB15K237, we use the following parameters: Batch size of 128, number of epochs of 500, learning rate of 0.0005, decay rate of 1.0, embedding dimension of 200, input dropout of 0.3, hidden dropout of 0.4 and 0.5, and label smoothing of 0.1. For YAGO15K, we use the following parameters: Batch size of 128, number of epochs of 500, learning rate of 0.003, decay rate of 0.99, embedding dimension of 200, input dropout of 0.2, hidden dropout of 0.2 and 0.3, and no label smoothing. The use of different parameters is indicated in \cite{balavzevic2019tucker}.

For experiments on Darpa Wikidata (DWD), we used the following parameters: embedding dimension of 200, batch size of 5000, number of negatives of 500, softmax loss function, learning rate of 0.1, relation learning rate of 0.01, regularization coefficient of 0.001, comparator of dot, and no eval fraction.

\begin{table*}[!t]
    \centering
    \small
    \caption{Link prediction results on FB15K-237. We include the originally reported results in \protect\cite{kristiadi2019incorporating} in parentheses. }
    \label{tab:link_prediction_reproducibility}
    \begin{tabular}{l l | l l l }
     & & \multicolumn{3}{c}{\bf FB15K-237} \\ 
      \bf embedding & \bf method & MRR & Hits@1 & Hits@10   \\ 
      \hline
      DistMult  & -             & 0.295 (0.282) & 0.212 (0.203) & 0.463 (0.438) \\
      DistMult  & LiteralE      & 0.309 (0.317) & 0.223 (0.232) & 0.481 (0.483) \\
      DistMult  & KBLN          & 0.302 (0.301) & 0.220 (0.215) & 0.470 (0.468) \\
      \hline
      ComplEx   & -             & 0.288 (0.290) & 0.205 (0.212) & 0.455 (0.445) \\
      ComplEx   & LiteralE      & 0.295 (0.305) & 0.212 (0.222) & 0.462 (0.466) \\
      \hline
      ConvE     & -             & 0.314 (0.313) & 0.226 (0.228) & 0.488 (0.479) \\
      ConvE     & LiteralE      & 0.317 (0.303) & 0.230 (0.219) & 0.489 (0.471) \\
    \end{tabular}

\end{table*}

\subsection*{Reproduced results}
As the codebase for implementing LiteralE and KBLN differs from the ones used in the original papers, we report both our obtained results as well as those from the original paper in Table~\ref{tab:link_prediction_reproducibility}. We have carefully compared our results against the results published in \cite{kristiadi2019incorporating}. The results of ours are mostly align theirs. For base models, DistMult tends to perform stronger, with MRR up by 0.013 and Hits@10 up by 0.025. For KBLN, the results are similar. For LiteralE, the results for DistMult and ComplEx are weaker, but Hits@10 is close. For ConvE, we have observed stronger than baseline results, unlike what is reported in \cite{kristiadi2019incorporating}. 

\begin{table*}[!t]
    \centering
    \small
    \caption{Performance of our numeric predictor with different choices of base model on graph augmented with QOC, when compared to existing SOTA methods on the FB15K-237 and YAGO15K dataset. Numbers indicate MAE. 
    Values of NAP++ and MrAP are taken from~\protect\cite{bayram2021node}.} 
    \label{tab:value_imputation}
    \begin{tabular}{l | l r r r r r r r r }
    & & \multicolumn{6}{c}{\bf KGA} & \bf NAP++ & \bf MrAP \\
     \bf dataset & \multicolumn{1}{c}{\bf attribute} & \bf TransE & \bf DistMult & \bf ComplEx & \bf ConvE  & \bf RotatE & \bf TuckER \\ 
     \hline
     \multirow{11}{*}{FB15K-237} 
     & date\_of\_birth       & 18.9     & 24.0  & 23.3  &   21.3    &   19.1    &   18.1    &   22.1      & \bf 15.0    \\
     & date\_of\_death       & 20.6     & 24.5  & 29.2  &   17.8    &   17.4    &   19.1    &   52.3      & \bf 16.3    \\
     & film\_release         & 4.0      & 5.2   & 4.7   &   3.4     &   \bf 3.3     &   3.6     &   9.9       & 6.3     \\
     & organization\_founded & \bf 49.0     & 63.9  & 63.1  &   57.1    &   54.4    &   56.1    &   59.3      & 58.3    \\
     & location\_founded     & \bf 76.0     & 95.6  & 129.7 &   94.8    &   79.2    &   89.3    &   92.1      & 98.8    \\
     & latitude              & 2.1      & 6.1   & 5.6   &   4.0     &   4.1     &   2.8     &   11.8      & \bf 1.5     \\
     & longitude             & 7.1      & 19.2  & 13.4  &   7.6     &   8.8     &   6.3     &   54.7      & \bf 4.0     \\
     & area                  & 6.1e4    & 1.1e5 & 7.4e4 &   \bf 3.6e4   &   6.6e4   &   8.0e4   &   4.4e5     & 4.4e5   \\
     & population            & 4.0e6    & 3.9e6 & 3.7e6 &   3.5e6   &   3.8e6   &   \bf 2.9e6   &   7.5e6     & 2.1e7   \\
     & height                & 0.077    & 0.078 & 0.074 &   0.072   &   0.070   &   \bf 0.069   &   0.080     & 0.086   \\
     & weight                & 11.6     & 13.4  & 10.7  &   9.5     &   10.5    &   \bf 8.0     &   15.3      & 12.9    \\ 
     \hline
     \multirow{7}{*}{YAGO15K} 
     & date\_of\_birth       & \bf 16.3     & 18.4     & 23.0  &   18.9    &    17.6    &   20.6    & 23.2 & 19.7 \\
     & date\_of\_death       & 30.8     & 35.8     & 38.7  &   31.9    &    35.8    &   \bf 30.6    & 45.7 & 34.0 \\
     & date\_created         & \bf 58.2     & 84.0     & 104.6 &   88.2    &    60.4    &   81.7    & 83.5 & 70.4 \\
     & data\_destroyed       & 23.3     & 31.1     & 25.6  &   27.5    &    \bf 22.8    &   26.8    & 38.2 & 34.6 \\
     & date\_happened        & \bf 29.9     & \bf 29.9     & 34.3  &    30.7    &    38.5    &   52.8    & 73.7 & 54.1 \\
     & latitude              & 3.4      & 9.9      & 10.4  &   6.1     &    4.9     &   7.2     & 8.7  & \bf 2.8 \\
     & longitude             & 7.2      & 23.7     & 27.2  &   11.7    &    11.1    &   10.8    & 43.1 & \bf 5.7 \\ 
     \hline
    \end{tabular}
\end{table*}

\subsection*{Model Parameters}

Comparison of the model complexity in terms of their parameters is given in Table~\ref{tab:params}. 
When compared with LiteralE and KBLRN, KGA may potentially have more parameters to estimate per training epoch. However, instead of introducing more complexity during the forward and backward passes, KGA reduces the complexity by not modifying the original embedding model. KGA shines when the base model is simple (such as TransE and DistMult) and the cost for estimating new parameters is smaller than estimating GRU for LiteralE and the RBF operation for KBLN.

\begin{table}[!t]
    \centering
    \small
    \caption{Comparison of method parameters, adapted from~\protect\cite{gesese2019survey}. $\Theta$ is the number of parameters in the base model, $H$ is the entity embedding size, $N_d$ is the number of attributes, $\Lambda$ is the size of the hidden layer in the Attrnet networks of MTKGNN, $N_r$ is the number of relations, $M$ is attribute embedding size, $b$ is the number of bins.} 
    \label{tab:params}
        \begin{tabular}{l | l}
        \multicolumn{1}{l|}{Model}  & \#Parameters   \\
        \hline
        TransEA                           & $\Theta + N_dM$                                   \\
        MTKGNN                            & $\Theta+NdH+2(2H\Lambda+\Lambda)$                 \\
        LiteralE                          & $\Theta+2H^2+2N_dH+H$                             \\
        KBLRN                             & $\Theta+N_rN_d$                                   \\ \hline
        \bf Our Model                     & $\Theta + N_dM + bH$                              \\ \hline
        \end{tabular}

\end{table}

\subsection*{Distribution of optimal bin sizes and discretization strategies}

Table~\ref{tab:link_prediction} shows the best result across discretization strategies and bin sizes. Here we report on the configuration chosen for each of these results.

For FB15K237, QSC (quantile, single, overlapping) performs best with TuckER, while QHC (quantile, hierarchy, chaining) works best for the 5 remaining models. Regarding bin sizes, 16 bins works best with TransE and RotatE, while 32 bins works best with the four remaining models. 

For YAGO15K, QHC is the best binning strategy for 5 of the 6 models. QOC performs best with ComplEx. 16 bins yields the best result in 4 of the 6 models, 32 bins works best with TuckER, and 8 bins work best with ConvE. 

\subsection*{Extended numeric link prediction results}

We have included Table \ref{tab:value_imputation} for users to compare the performance of value imputation of different embedding models in the two datasets. In general, models with stronger link prediction performance tends to have better value imputation performance. TuckER tends to beat other models for FB15K237, and ConvE is a competitive model for YAGO15K. It is a bit surprising that TransE performs very well in value imputation, when compared with its link prediction results. It could be caused by the fact that TransE works well with many-to-one relations, which is the typical relationship between entities and their attributes.

\end{document}